\pdfoutput=1

\documentclass[11pt]{article}

\usepackage[]{eacl2023}

\usepackage{times}
\usepackage{calc}
\usepackage{graphicx}
\usepackage{tabularx}
\usepackage{multirow}
\usepackage{booktabs}
\usepackage{latexsym}
\usepackage{amsmath}
\usepackage{algorithm}
\usepackage{algpseudocode}
\usepackage{bbm}
\usepackage{pifont}

\definecolor{greenrgb}{rgb}{0.18, 0.71, 0.18}
\newcommand{\cmark}{\ding{51}}%
\newcommand{\xmark}{\ding{55}}%

\usepackage{booktabs}
\usepackage[T1]{fontenc}

\usepackage[utf8]{inputenc}

\usepackage{microtype}

\usepackage{xcolor}
\usepackage[T2A, T1]{fontenc}
\usepackage[russian, english]{babel}
\usepackage{substitutefont}
\PassOptionsToPackage{hyphens}{url}\usepackage{hyperref}
\usepackage{xurl}

\usepackage[normalem]{ulem}

\newcommand{\example}[1]{\emph{``#1''}}

\newcommand{\winox}[0]{Wino-X}
\newcommand{\winomt}[0]{WinoMT}
\newcommand{\bug}[0]{BUG}

\title{Evaluating and Improving the Coreference Capabilities of \\ Machine Translation Models}

\author{Asaf Yehudai\textsuperscript{1} \quad 
        Arie Cattan\textsuperscript{2} \quad
        Omri Abend\textsuperscript{1} \quad 
        Gabriel Stanovsky\textsuperscript{1}\\ 
        \textsuperscript{1}School of Computer Science, The Hebrew University of Jerusalem \\ 
        \textsuperscript{2}Computer Science Department, Bar Ilan University  
        \\ 
  {\normalsize\tt  \{asaf.yehudai,omri.abend,gabriel.stanovsky\}@mail.huji.ac.il} \\
  \normalsize\tt arie.cattan@gmail.com \\
 }
\begin{document}
\maketitle

\begin{abstract}

Machine translation (MT) requires a wide range of linguistic capabilities, which current end-to-end models are expected to learn implicitly by observing aligned sentences in bilingual corpora.
In this work, we ask: \emph{How well do MT models learn coreference resolution from implicit signal?} To answer this question, we develop an evaluation methodology that derives coreference clusters from MT output and evaluates them without requiring  annotations in the target language.
We further evaluate several prominent open-source and commercial MT systems, translating from English to six target languages, and compare them to state-of-the-art coreference resolvers on three challenging benchmarks.
Our results show that the monolingual resolvers greatly outperform MT models. Motivated by this result, we experiment with different methods for incorporating the output of coreference resolution models in MT, showing improvement over strong baselines.\footnote{\url{https://github.com/AsafYehudai/MT-coref}}

\end{abstract}
\section{Introduction}
\label{sec:intro}

Machine translation (MT) may require coreference resolution to translate cases where the source and target language differ in their grammatical properties. For example, consider translating \example{The trophy didn't fit in the suitcase because it was too small} from English to French: \example{Le trophée ne rentrait pas dans la valise car elle était trop petite} \citep{Sakaguchi2020WINOGRANDEAA}. In French, suitcase~(\example{valise}) is grammatically feminine, and trophy~(\example{trophée}) is masculine, while the source-side English does not encode grammatical noun gender. This requires an MT model to infer that \example{it} refers to the suitcase (and not the trophy) to correctly produce the feminine inflection for the phrase \example{it was too small} (\example{elle était trop petite}), whereas an incorrect coreference resolution may produce the masculine inflection (\example{il était trop petit}) corresponding to the trophy.

\begin{figure}[t]
\centering
\includegraphics[scale=0.38]{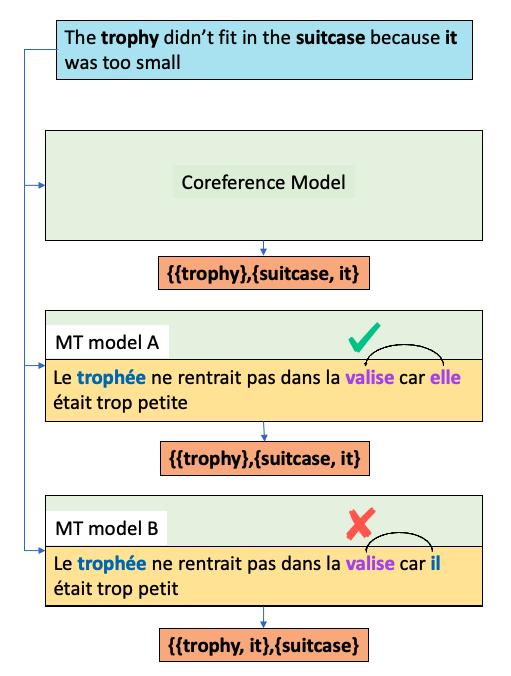}
\caption{MT models can be compared to source-side coreference resolvers. An example translation from English (turquoise) to French (yellow). Our method first identifies the grammatical gender of the mentions in the target language marked in purple (female) and blue (male), followed by inferring the source side clusters (orange), through gender agreement.}
\label{fig:method_new}
\end{figure}


Such texts evade lexical one-to-one translation, and instead demand source-side coreference resolution as a prerequisite for a correct translation. 
The prominent end-to-end approach to MT assumes that translation models implicitly learn source-side coreference resolution by observing aligned source-target pairs, without intermediate coreference supervision. While the importance of addressing such semantic phenomenon has been stated in various works~\citep{le-nagard-koehn-2010-aiding,stojanovski-fraser-2018-coreference}, it was also observed that the ubiquitous BLEU metric~\citep{bleu} does not adequately quantify it~\citep{hardmeier-federico-2010-modelling, freitag2022results}.




This work addresses the following research question: \textit{How well does MT learn coreference when compared against explicit coreference supervision?}
Answering this question can improve our understanding of the way MT models operate and 
also has practical implications: if implicit supervision lags behind monolingual training, it would motivate integration between end-to-end MT approaches and explicitly-supervised monolingual components.

In Section~\ref{sec:consistency_metric} we devise an evaluation paradigm that reduces MT output to source-side coreference resolution predictions by inferring coreference clusters from source inputs and predicted target translations. E.g., in the previous example, 
a feminine inflection for the pronoun ``it'' in French can infer linking ``it'' with ``suitcase'', while a masculine French inflection links ``it'' with ``trophy'', as shown in Figure~\ref{fig:method_new}.
This approach allows us to distill the coreference resolution abilities of MT models and compare them against state-of-the-art coreference resolution models, trained explicitly on the task.

We use this approach to evaluate the coreference capabilities of several commercial and open source MT systems, translating from English to six target languages. We conduct our experiments in both synthetic \citep[\winomt{} and \winox{};][]{stanovsky-etal-2019-evaluating,emelin-sennrich-2021-wino}
and naturalistic settings  \citep[\bug{};][]{levy-etal-2021-collecting-large}.
Our results show that state-of-the-art coreference resolvers vastly outperform MT models on several benchmarks, indicating that explicit supervision may lead to better coreference performance.

Following this finding, in Section~\ref{sec:improving}, we develop methods for improving coreference in MT, both implicitly and explicitly.
Our implicit approach consists of fine-tuning MT models on 
texts that specifically require many coreference decisions, thus exposing the model to more implicit coreference signal.
Our explicit approach further enriches source sentences with predicted coreference markers. 
We show that these approaches improve coreference over the end-to-end MT approach, achieving comparable or better results than much larger MT models, both commercial systems and open-source.

More broadly, our approach can be applied to improve the translation of other semantic phenomena that diverge in realization between source and target languages, such as plurality in second-person pronouns~\citep{Stanovsky2019YallSR} or tense marking~\citep{wolfram1985variability}. 

\section{Background: Gender Bias in MT}
\label{sec:background}

We start our work by extending the methodology developed in ~\citep{stanovsky-etal-2019-evaluating}, which relies on target-side morphology to infer the translated gender of certain professions.

In particular, assuming a dataset of English sentences $D$, where each instance includes gold coreference annotation between a human entity and its pronoun (e.g., \example{The \textbf{doctor} asked the nurse to help \textbf{her} with the procedure.}), they evaluate gender bias from English to language $T$ with morphological gender in the following manner:

\begin{enumerate}
    \item Predict word alignment between $D$ and $M(D)$, i.e., the output translations of an MT model $M$. This finds the translations for pronouns (e.g., \example{her}) and possible entities (e.g \example{doctor}, \example{nurse}) in the target language $T$.
    \item Automatically extract the gender of the possible entities and the pronouns in the target language based on morphological features.
    \item Check whether the gender of the co-referred entity (e.g., \example{doctor}) in $T$ corresponds to the gender of the English pronoun (e.g., \example{her}).
\end{enumerate}

The gender bias of $M$ is then defined as the difference in performance between stereotypical and anti-stereotypical gender role assignments. 

We use a similar setup to address a different question: rather than evaluating the gender bias of the model, we evaluate its coreference abilities, which may be hindered by bias, but also by the inherent difficulty to infer coreference in the absence of an explicit training signal.

\section{MT Models Fare Poorly Against Coreference Resolvers}
\label{sec:consistency_metric}

The approach taken in \winomt{} is limited as it restricts the evaluation to sentences with a known gender in English, indicated by a gendered pronoun of a human entity (e.g., \emph{her}). Consider the sentence in Figure~\ref{fig:method_new}: \example{The trophy didn't fit in the suitcase because it was too small} with the coreference cluster \{suitcase, it\}. Step 3 in \citeauthor{stanovsky-etal-2019-evaluating}'s method will fail to assign a coreference label to the translation, because \example{it} does not have a gender in English. In this section, we extend the \winomt{} approach in order to estimate more general coreference abilities of MT models.

To achieve this, we note that many languages have gender agreement between pronouns and the noun that they refer to. Therefore, correct target-side gender agreement requires (implicitly) resolving the source-side coreference of the relevant entities. As exemplified in Figure~\ref{fig:method_new}, a reader of the French translation 
would infer from the gender inflection of ``it'' whether it refers to ``suitcase'' or the ``trophy''. I.e., a feminine pronoun (``elle'') would agree with the feminine noun suitcase (``valise''), while a masculine pronoun (``il'') would agree with the masculine noun trophy (``trophée'). 
We therefore formulate a new metric quantifying the ability of the MT model to implicitly resolve source-side coreference (henceforth,  \textit{Target-side Consistency}), defined as the proportion of instances in which the morphological gender of an entity (e.g., ``suitcase'') matches that of its referring pronoun (e.g., ``it'') in the target language $T$.

This metric examines whether an MT model is consistent in its coreference decisions, regardless of whether it correctly inferred the coreference relations in the input text. Indeed, some texts may keep the English ambiguity in the translation, and hence absolve the MT model from resolving coreference. For example, in the sentence \example{The battery didn't fit in the suitcase because it was too small}, both \emph{battery} and \emph{suitcase} are feminine in French. Our proposed metric will correctly indicate that the MT model was successful in such cases (albeit trivially). The metric thus serves as an upper bound on the MT model's coreference abilities.

We note that while this framing uses the morphological gender inflection of common nouns, it is different in motivation from measures of gender bias. In our example above, gender inflection allows us to determine whether an MT model correctly employs common sense rather than examining whether it tends to prefer stereotypical gender norms. While a model's gender bias may explain some loss in coreference abilities, the model's ability to resolve coreference need not be aligned with the degree of its bias (e.g., a random gender assignment would result in
unbiased performance, but very poor coreference ability).

Most importantly, by considering the gender of the entity and the pronoun, we obtain mention clusters which can be compared against those produced by coreference resolution models. In our example figure, both the first MT model and the coreference model produce the correct clustering: \{\{trophy\}, \{suitcase, it\}\}, while the second MT model errs by producing: \{\{trophy, it\}, \{suitcase\}\}.


Another aspect of our evaluation methodology is its generality. Our method does not require a reference translation or make any particular assumptions about the generated output.  As there are generally many correct translations, this flexibility allows us to accurately assess the model's coreference abilities. For instance, our methodology does not assume the gender of the entity's translation as can be seen in the first example in Table \ref{table:lang_exampels} where the two systems translate the entity ``jar'' differently. Further, some languages might not always translate an English pronoun into a pronoun but still express its gender in a different word. Consider the second example in Table~\ref{table:lang_exampels} where the alignment model (step 2 in §\ref{sec:background}) finds that the English word ``it'' is aligned with both ``l'' and ``trouvée'' in French. Here, the feminine suffix of the past participle ``trouv\textbf{ée}'' indicates that the ellipsis ``l'' corresponds to a feminine entity. 



\subsection{Evaluation Setup}
\label{subsec:eval}

\paragraph{Evaluation datasets.} 
The first dataset we use is \winox{}~\citep{emelin-sennrich-2021-wino}, a filtered subset of WinoGrande \citep{Sakaguchi2020WINOGRANDEAA} built to test commonsense reasoning and coreference resolution of MT models and multilingual encoders. The dataset contains sentences similar to the one in Figure~\ref{fig:method_new}. All sentences have two entities and a pronoun, ``it'', coreferring to one of them. The dataset consists of three parts, each part constructed for a different target language (German, French, and Russian), where each part only contains sentences where the two entities have different morphological gender in the target language (e.g., in French \example{trophy} is masculine whereas \example{suitcase} is feminine). Hence, applying our target-side consistency metric on these filtered sentences avoids trivial instances where both candidate entities have the same gender in the target language, and provides a clearer picture of the coreference capabilities of the model.

Second, we use \winomt{}~\citep{stanovsky-etal-2019-evaluating},\footnote{\winomt{} is a combination of Winogender~\citep{rudinger-etal-2018-gender} and WinoBias~\citep{zhao-etal-2018-gender}.} a dataset built following the Winograd schema \citep{Levesque2011TheWS}, designed to test gender bias and coreference resolution of MT models. The sentences in this dataset contain two human entities and one gendered pronoun, e.g., \example{The doctor asked the nurse to help her in the procedure}. The gendered pronoun reveals the gender of the entity and adds gender attributes to the source cluster. In our example, \example{her} refers to the \example{doctor}, revealing the doctor's gender. 

Our third dataset is \bug{}~\citep{levy-etal-2021-collecting-large}, a semi-automatic collection of naturalistic English sentences that are challenging with respect to societal gender-role assignments. Similar to WinoMT, each sentence contains a human entity, identified by their profession and a gendered pronoun. 
To reduce noise, we use the \textsc{Gold} portion of this dataset which was validated by human annotators. All datasets statistics are presented in Table \ref{table:data_val}.

\begin{table}[t]
\centering
\begin{tabular}{lr}
\toprule
Dataset  & \#sentences \\
\midrule
Wino-X (en $\rightarrow$ de) & 3,774 \\
Wino-X (en $\rightarrow$ fr) & 2,988 \\
Wino-X (en $\rightarrow$ ru) & 2,238 \\
WinoMT      & 3,888 \\
BUG         & 1,717 \\
\bottomrule
\end{tabular}
\caption{Statistics of our evaluation datasets. Note that in all datasets we use the corresponding English source-side sentences as our input.}
\label{table:data_val}
\end{table}

\paragraph{Machine translation models.} We apply our evaluation methodology to four Transformer-based machine translation models from EasyNMT:\footnote{https://github.com/UKPLab/EasyNMT} mBART50 \citep{https://doi.org/10.48550/arxiv.2008.00401, liu-etal-2020-multilingual-denoising}, M2M\_418M, M2M\_1.2B \citep{Fan2021BeyondEM}, and the bilingual
Opus-MT~\citep{tiedemann-thottingal-2020-opus}, representing the state-of-the-art for publicly available neural machine translations models.
In addition, we measure coreference consistency on the output of two commercial systems: Google Translate\footnote{https://cloud.google.com/translate} and Microsoft Translator.\footnote{https://www.bing.com/translator}

\paragraph{Target languages.} For WinoMT and BUG, we translate from English to six different languages: Arabic, German, Spanish, Hebrew, Russian and French. These languages form a diverse set with respect to how they encode grammatical gender (e.g., number of grammatical genders), as well as to their orthography, word order and other linguistic traits, while still allowing for highly accurate automatic morphological analysis.
These languages belong to four families: (1) Romance languages: Spanish and French, which have gendered noun-determiner agreement with two grammatical genders; Spanish is also a \emph{pro-drop} language, i.e., pronouns can be omitted in certain cases, which in our setting may keep the coreference ambiguity of the source-side English sentence~\citep{Webster2020ScalableCL}.  
(2) Slavic languages (with Cyrillic alphabets): Russian with 3 grammatical genders. (3) Semitic languages: Hebrew and Arabic, each with a unique alphabet; both are partial pro-drop languages and have two grammatical genders. (4) Germanic languages: German with 3 grammatical genders.

\subsection{Target-side Consistency Results}
\label{subsec:evaluate_cons}

We first evaluate the accuracy of existing coreference resolvers on our three evaluation datasets, where accuracy is defined as the percentage of instances in which the model identifies that the pronoun is coreferring with the correct entity.
We select state-of-the-art models trained on CoNLL-2012~\citep{pradhan-etal-2012-conll}: SpanBERT~\citep{Joshi2020SpanBERTIP},\footnote{Using AllenNLP's implementation~\citep{Gardner2018AllenNLPAD}.} the s2e model~\citep{Kirstain2021CoreferenceRW} and \textsc{LingMess}~\citep{Otmazgin2022LingMessLI}.
Results in Table \ref{table:coref_models} show that coreference models perform quite well on \winomt{} and \bug{} but poorly on \winox{} (60.8 for \emph{s2e}), indicating weak commonsense capabilities. 

\begin{table}[tb!]
\centering
\resizebox{0.4\textwidth}{!}{
\begin{tabular}{@{}lcccc@{}}
\toprule
     & Wino-X & WinoMT & BUG \\
\midrule
SpanBERT            & 51.2     & 76.6 &  72.0 &  \\
\emph{s2e}          & \textbf{60.8}     & 81.7 &  72.2 &  \\
\textsc{LingMess}  & 58.7     & \textbf{83.7} &  \textbf{74.6} &  \\
\bottomrule
\end{tabular}}
\caption{Accuracy of SpanBERT, \emph{s2e} and \textsc{LingMess} model on our evaluation datasets. For simplicity, we report the accuracy on Wino-X sentences from the three languages as a single corpus, because there is a small difference between the languages (up to 0.3). }
\label{table:coref_models}
\end{table}
\begin{table}
    \centering
    \begin{tabular}{lccc}
    \toprule
    & en $\rightarrow$ de & en $\rightarrow$ fr & en $\rightarrow$ ru \\
    \midrule
    mBART50   & 37.4         & \textbf{56.6} & 44.6    \\
    M2M\_418M & 31.4         & 48.5          & \textbf{44.9} \\
    Google         & \textbf{41.3}& 36.3          & 40.7 \\
    Microsoft      & 40.5         & 36.7          & 43.5 \\
    Opus-MT       & 37.4         & 35.3          & 43.6 \\
    \bottomrule
    \end{tabular}
    \caption{Target-side consistency results of commercial and open-source MT systems on Wino-X when translating into German, French, and Russian.}
    \label{tab:winox_cons}
\end{table}

\begin{table*}[!htb]
    \centering
    \scalebox{0.95}{
           \begin{tabular}{llcccccclcccccc}
            \toprule
            && \multicolumn{6}{c}{WinoMT} && \multicolumn{6}{c}{BUG} \\
            \midrule
             
             &&
             de & 
             es & 
             fr & 
             ru & 
             he & 
             ar &&
             de & 
             es & 
             fr & 
             ru & 
             he & 
             ar \\
             
            \midrule
            mBART50   && 77.7 & \textbf{75.2} & \textbf{73.3} & 57.7 & \textbf{69.3} & \textbf{69.5} && 74.6 & 69.4 & 76.0 & 69.1 & 80.2 & 83.0 \\
            M2M\_418M && 69.7 & 55.1 & 65.4 & 54.7 & 56.5 & 64.3 && 81.3 & 84.9 & 86.0 & \textbf{71.6} & 83.6 & 85.1  \\
            M2M\_1.2B && 69.7 & 55.1 & 65.4 & 54.7 & 56.5 & 64.3 && \textbf{83.5} & 85.1 & \textbf{90.3} & 70.4 & \textbf{89.6} & \textbf{93.4} \\
            Google   && 69.9 & 59.4 & 65.5 & \textbf{57.9} & 60.1 & 69.2 && 56.0 & 84.4 & 89.2 & 71.0 & 85.2 & 91.9 \\
            Microsoft     && \textbf{78.0}   & 66.5 & 69.3 & 57.3 & 63.9 & 60.7 && 77.2 & \textbf{86.7} & 86.0 & 67.2 & 84.0 & 89.8  \\
            Opus-MT       && 68.2 & 56.0   & 63.2 & 49.8 & 59.0   & 59.8 && 79.8 & 85.1 & 86.2 & 67.3 & 88.8 & 88.0\\
            \bottomrule
    \end{tabular}}
    \caption{Target-side consistency results of commercial and open-source MT systems on WinoMT and BUG when translating into different languages. These numbers are an upper bound for the source-side coreference accuracy.\vspace{-.3cm}}
    \label{table:winomt_bug}
\end{table*}

Table~\ref{tab:winox_cons} shows the target-side coreference consistency scores for all MT models on \winox{}, which, as mentioned above~(§\ref{subsec:eval}), includes only sentences where the entity and pronoun should be translated using different genders in the target language. We observe that all MT models perform poorly on \winox{} with the highest average score of 46.2 for mBART50, which vastly underperforms English coreference resolvers by 14.6 points. Interestingly, many instances are inconsistent because models tend to generate \textit{neutral} pronoun whereas a gendered pronoun is expected, for example,  {\it cela}, {\it c'était} in French ($31\%$ of translations) and \foreignlanguage{russian}{это,они} in Russian ($17\%$ of translations), meaning ``this'' or ``they'' in English. Likewise, $68\%$ of German translations include neutral pronouns (e.g., ``es''), while only $22\%$ of the entities are neutral. The reported percentages were calculated on Opus-MT. Similar trends were observed in all models.

Table~\ref{table:winomt_bug} shows the target-side consistency results for \winomt{} and \bug{}. Following common practices on those datasets~\citep{stanovsky-etal-2019-evaluating, levy-etal-2021-collecting-large}, we omit sentences where the candidate pronoun does not provide information about the entity's gender. For example, in French, possessive pronouns agree with the gender of the possessed object, rather than the possessor as in English. Another example is in Spanish, which is a pro-drop language, where a valid translation can drop the pronoun and use a generic verb, leaving the only gender signal in the translation to be marked on the profession noun. See App. \S\ref{sec:appendix:04_models_exampels} for more examples.

Similarly to \winox{}, target-side consistency results on \winomt{} are consistently lower than coreference resolvers. Further, we observe that consistency is affected by two factors: the MT model and the target language. Regarding models, Opus-MT achieves lowest performance, with average consistency of $59.3$, while mBART50 achieves high results with average consistency of $70.5$, sometimes surpassing the second-best MT model by about 9 points. This might be due to the extensive pre-training of mBART50, as previously demonstrated for monolingual LMs \citep{Huang2019CosmosQM, Sakaguchi2020WINOGRANDEAA, Bhagavatula2020AbductiveCR}. With respect to target languages, Russian consistency results are systematically lower than the results in other languages, to the extent that the best model in Russian provides lower results than the worst model in most other languages. In contrast, all models in German achieve a consistency score of about $70$ or more, which can be due to its similarity with English and the research focus on improving English-German translations. 

Consistency results on BUG are higher than on WinoMT for most models, while sometimes surpassing English coreference resolvers, notably in Hebrew and Arabic (e.g., 91.8 for Google vs. 74.6 for \textsc{LingMess}). To understand this gap, we analyze the translation of 50 BUG sentences to Hebrew and French and find that most instances (45 in Hebrew and 33 in French) do not include a distracting entity which should be translated to a different gender in the target language. As mentioned above~(§\ref{sec:consistency_metric}), our metric trivially indicates those examples as consistent.

Overall, target-side consistency results across all datasets demonstrate that both open-source and commercial MT systems exhibit rather poor coreference capabilities compared to English coreference models.

\subsection{Human Validation}

The use of automatic tools in the proposed methodology inevitably implies the introduction of noise into the process.
To assess the quality of our measurements, we randomly sampled 50 translations of the Opus-MT model from all evaluation datasets and in all target languages (for a total of 750 annotations), annotating each sample in-house by a native speaker of the target language. The human annotators were asked to identify if the candidate pronoun is indeed the target pronoun and to verify that the gender prediction is correct. This way, we can account for both types of possible errors, i.e., alignment and gender extraction. 

We compare the human annotations to the output of our automatic method and find that the average agreement over all languages and datasets is above $90\%$ (see full results in App. \S\ref{sec:appendix:05_human_validation_results}). These results are comparable to the ones reported by \citet{stanovsky-etal-2019-evaluating}, who conducted human validation and reported that their alignment and gender prediction of the entity in question were reliable for $85\%$ of translations across all languages. 

Some errors can be caused by idiosyncrasies that affect the morphological analysis, as \citet{stanovsky-etal-2019-evaluating} noted.
For example, gender for certain words in Hebrew cannot be determined without diacritics, and some pronouns in German are used in both masculine and neutral forms (e.g., sein), or feminine and third-person plural forms (e.g., ihr). In addition, we notice that sentences from \bug{}, specifically in partial pro-drop languages, were found to be more challenging for the alignment model, and account for most mistakes in Hebrew and Arabic.

\section{Improving MT Coreference Consistency}
\label{sec:improving}

In the previous section we showed that the coreference performance of MT systems, obtained through an implicit signal, seems inferior to that of coreference resolution learned from an explicit signal. This result raises the question of whether we can leverage dedicated conference resolvers to improve the consistency of MT coreference.

To address this question, we propose two data augmentation techniques that leverage a source-side English coreference model, and show that finetuning on them indeed improves coreference resolution in MT.

\paragraph{Augmented fine-tuning with instances which require coreference resolution.} First, we run a coreference resolution model on the source-side sentences. We then consider two approaches for constructing the augmented fine-tuning data: (1) \emph{Coref data} with all sentences that have non-singleton clusters and (2) \emph{Gender data}, a subset of \emph{Coref data} where there is at least one non-singleton cluster with a gendered pronoun (he, she, her, him, hers, his). The motivation for this augmented fine-tuning strategy is that further fine-tuning on such instances would expose the MT model to examples that may bear a coreference signal. 

\paragraph{Adding explicit source-side coreference markers.}
Second, we use the non-singleton clusters from the coreference model to add inline coreference markers in the source sentences. For our example sentence, this process produces the following source-side sequence: \emph{``The trophy didn't fit in the \textbf{<ENT1>} suitcase \textbf{</ENT1>} because \textbf{<ENT1>} it \textbf{</ENT1>} was too small''}, indicating that ``suitcase'' and ``it'' are coreferring. 

\subsection{Experimental Setup}
\label{subsec:improve_exp}

\paragraph{MT models.} In our fine-tuning experiments, we opt for the Opus-MT model, since its size (68M parameters) and efficiency \citep{junczys-dowmunt-etal-2018-marian} enables us to run extensive experiments across many languages.

\paragraph{Training datasets.}
For fine-tuning data of Spanish, French, and German, we use Europarl \citep{Koehn2005EuroparlAP}, and for Russian, Hebrew, and Arabic, we use CCMatrix \citep{Schwenk2021CCMatrixMB}, randomly sub-sampled to 5M sentences for computational reasons. In each dataset, we find instances that require coreference resolution and add appropriate markup using the \emph{s2e} coreference resolver. We use \emph{s2e} as it performed well in the previous experiments. Table \ref{table:data_train} shows the size of \emph{Coref data} and \emph{Gender data} for all training datasets. Note that invariably \emph{Gender data} is an order of magnitude smaller than \emph{Coref data}.

\begin{table}[t]
\centering
\resizebox{\columnwidth}{!}{
\begin{tabular}{lllllll}
\toprule
            & de   & es   & fr   & ru   & he   & ar   \\
\midrule
Coref data  & 500K & 744K & 761K & 149K & 1.1M & 1.4M \\
Gender data & 38K  & 50K  & 51K  & 19K  & 265K & 268K \\
\bottomrule
\end{tabular}}
\caption{Number of fine-tuning instances in \emph{Coref Data} (requiring some sort of coreference resolution) and \emph{Gender data} (requiring coreference resolution with some gendered pronoun) for each target language.}
\label{table:data_train}
\end{table}

\paragraph{Fine-tuning and inference.}
For each language, we fine-tune the Opus-MT model using four different finetuning datasets: (1) \emph{Coref data} (2) \emph{Coref data} with explicit coreference markers, (3) \emph{Gender data} and (4) \emph{Gender data} with explicit coreference markers. The inference on our three evaluation datasets (Wino-X, WinoMT, BUG) conforms with the fine-tuning procedure of each model. Namely, we run the models (1) and (3) on raw English sentences. For the models (2) and (4), we first add explicit coreference markers according to the output of the \emph{s2e} model (a) or the gold annotation (b), then translate those augmented sentences to the different languages. 

\subsection{Results}
\label{sec:results}

\begin{table*}[!htb]
    \centering
    \scalebox{0.9}{
           \begin{tabular}{llccc|cccccc}
            \toprule
            && \multicolumn{3}{c}{Wino-X} & \multicolumn{6}{c}{WinoMT} \\
            \midrule
             && en$\rightarrow$de        & en$\rightarrow$fr        & en$\rightarrow$ru &
              \multicolumn{1}{c}{en$\rightarrow$de} &
              \multicolumn{1}{c}{en$\rightarrow$es} &
              \multicolumn{1}{l}{en$\rightarrow$fr} &
              \multicolumn{1}{l}{en$\rightarrow$ru} &
              \multicolumn{1}{c}{en$\rightarrow$he} &
              \multicolumn{1}{c}{en$\rightarrow$ar} \\
            \midrule
                Opus-MT      && 37.4& 35.3& 43.6 & 67.6 & 56.0   & 63.2 & 49.8 & 59.0   & 59.8 \\
                Coref data (1)    && 42.7& 41.0& 44.0 & 74.4 & 58.0 & 67.1 & \textbf{62.1} & 68.1 & 67.1 \\
                $+$coref markers (2a)   && 44.1& 42.6& 45.5 & 76.0 & 58.2 & 67.2 & 57.7 & 68.7   & 67.9 \\
                $+$gold markers (2b)  && \textbf{44.7} & \textbf{45.8}& \textbf{50.5}   & 77.6 & 58.2 & 67.3 & 58.6 & 69.8   & \textbf{68.9} \\
                Gender data (3)   && 41.2&37.0&45.0 & 75.0 &60.6& 68.3 & 61.6 & 68.1 & 66.2 \\
                $+$coref markers (4a)      && 43.6 & 36.6 & 43.8 &  78.7 & 60.1 & \textbf{68.6} & 58.4 & 69.0   & 60.9 \\
                $+$gold markers (4b)       && 42.9 & 36.7 & 46.3 & \textbf{80.6} & \textbf{60.8} & 68.4 & 59.6 & \textbf{69.9}   & 62.2 \\
            \bottomrule
        \end{tabular}}
    \caption{Target-side consistency results of the Opus-MT baseline and our fine-tuning experiments on Wino-X and WinoMT when translating into different languages. For both datasets, our fine-tuned models surpass the baseline.}
    \label{table:cons_ours}
\end{table*}

Table~\ref{table:cons_ours} shows the target-side coreference consistency scores of all our fine-tuned models on \winox{} and \winomt{} (see App. \S\ref{sec:appendix:03_bug_results} for the performance on BUG, which follow similar trends). For both datasets, our fine-tuned models surpass the Opus-MT baseline model, while preserving the overall translation quality, as indicated by automatic measures such as BERTScore~\citep{bertscore} ($+0.08\%$) and COMET-20~\citep{rei-etal-2020-comet} ($+0.0025$).

\paragraph{Effect of augmented fine-tuning data.} The models fine-tuned on \emph{Coref data} (1) and \emph{Gender data} (3) outperform the Opus-MT baseline for all languages, both in Wino-X and WinoMT. This demonstrates that MT models learn implicitly linguistic phenomena from instances involving those phenomena. Furthermore, we point out that consistency scores on \winox{} are generally higher when fine-tuning on \emph{Coref data} (1, 2a, 2b) while WinoMT results are better when fine-tuning on \emph{Gender data} (3, 4a, 4b). This performance gap likely stems  from the similarity between WinoMT and \emph{Gender data} (as both include gendered pronouns), while Wino-X's like sentences with the pronoun ``it'' appear only in \emph{Coref data}. This further confirms the important role of fine-tuning data, which is in line with the observation of \citet{saunders-byrne-2020-reducing}, that smaller, more goal-oriented data is better for fine-tuning, compared to much larger but less focused data.

\paragraph{Effect of explicit coreference markers.} In the majority of our experiments (13/18), the explicit fine-tuning models (2a and 4a) outperform the implicit data augmentation approach when using the same augmented data (1 and 3) (see examples in Table \ref{table:lang_exampels}). These results suggest that an explicit monolingual signal can improve results more than an implicit signal. Results also show that the improvement is more pronounced when incorporating gold coreference markers (2b and 4b) instead of predicted markers (2a and 2b). Hence, applying more accurate coreference resolution models than the \emph{s2e} model will result in higher target-side consistency results.

\begin{table*}
    \centering
    \resizebox{\textwidth}{!}{
    \begin{tabular}{p{3cm}p{15cm}p{0.5cm}}
    \toprule


    \textbf{Source} & The chef tried to store the \textbf{fat} in the \textbf{jar} but \textbf{it} was too large. \\ 
    \textbf{Baseline (FR)} & Le chef a essayé de stocker \textcolor{red}{la graisse} dans \textcolor{blue}{le bocal}, mais \textcolor{blue}{il} était trop grand. & \textbf{\textcolor{red}{\xmark}} \\
    \textbf{Ours (FR)} & Le chef a essayé de stocker \textcolor{red}{la graisse} dans \textcolor{blue}{le pot}, mais \textcolor{red}{elle} était trop grande. & \textbf{\textcolor{greenrgb}{\cmark}} \\

    \midrule
    \textbf{Source} & The chickens escaped from the \textbf{yard} and fled to the \textbf{field}, as they found \textbf{it} so confining.  \\
         \textbf{Baseline (FR)} & Les poulets se sont échappés de \textcolor{red}{la cour} et ont fui vers \textcolor{blue}{le champ}, comme ils \textcolor{blue}{l'}ont trouv\textcolor{blue}{é} si restreint. & \textbf{\textcolor{red}{\xmark}} \\
         \textbf{Ours (FR)} & Les poulets se sont échappés de \textcolor{red}{la cour} et ont fui vers \textcolor{blue}{le champ}, car ils \textcolor{red}{l'}ont trouv\textcolor{red}{ée} si encombrée. &
  \textbf{\textcolor{greenrgb}{\cmark}} \\

  \midrule 
  \textbf{Source} & The headphones blocked the \textbf{noise} but not the \textbf{vibration}, as \textbf{it} was relatively strong \\ 
  \textbf{Baseline (RU)} & \foreignlanguage{russian}{Наушники блокировали \textcolor{blue}{шум}, но не \textcolor{red}{вибрацию}, поскольку \textcolor{blue}{он} был относительно сильным.} &
  \textbf{\textcolor{red}{\xmark}} \\
  \textbf{Ours (RU)} & \foreignlanguage{russian}{Наушники блокировали \textcolor{blue}{шум}, но не \textcolor{red}{вибрацию}, так как \textcolor{red}{она} была относительно сильной.} &
  \textbf{\textcolor{greenrgb}{\cmark}} \\
    \bottomrule
    \end{tabular}}
    \caption{Translation examples of Wino-X sentences to French and Russian by the baseline (Opus-MT) and our model (with coref markers). Words in blue and red indicate male, female entities, respectively. Bold indicates coreference mentions in the source sentence.}
    \label{table:lang_exampels}
\end{table*}

\subsection{Analysis}

We turn to observing the empirical effect of the suggested fine-tuning strategies, using additional metrics.
For each sentence in Wino-X, we have the gold target pronoun that should appear in its translation. We use it to compute pronoun translation accuracy by comparing the candidate pronoun with the gold target pronoun. Table \ref{table:ours_pm_winox} presents the results. We can see that our method provides a large improvement over the baseline. Comparing these results against those of prominent open-source and commercial MT (see App. \S\ref{sec:appendix:01_gmf_pm}) shows that our approach outperforms other MT models in German and Russian, and is only second in French.

In WinoMT, \citet{stanovsky-etal-2019-evaluating} computed {\it gender accuracy} as the percentage of instances in which the translation preserved the gender of the entity from the original English sentence~(§\ref{sec:background}). Table \ref{table:ours_acc_wino-mt} shows that our approach improved gender accuracy results across all languages except Arabic.

Other metrics that \citet{stanovsky-etal-2019-evaluating} used, are $\Delta S$ and $\Delta G$. $\Delta S$  measures the difference in gender accuracy between stereotypical and non-stereotypical gender role assignments (as defined by \citealp{zhao-etal-2017-men}), and $\Delta G$ measures the difference in performance (F1 score) between source sentences with male and female entities. Our method decreases biases in both $\Delta S$ and $\Delta G$ by 5-6 points on average, indicating that the explicit signal helps the model in associating the pronoun with the coreferring entity, even in the presence of social and gender biases.

\begin{table}[tb!]
\centering
\begin{tabular}{llll}
\toprule
            & en$\rightarrow$de        & en$\rightarrow$fr         & en$\rightarrow$ru        \\
\midrule
Opus-MT       & 39.8          & 31.7          & 37.0             \\
Coref data     & 42.3          & 38.9          & 35.0          \\
$+$coref markers    & 43.6          & 39.3          & 37.2          \\
$+$gold markers       & \textbf{45.7} & \textbf{42.7}           & \textbf{42.6} \\
\bottomrule
\end{tabular}
\caption{Pronoun accuracy results of our fine tuning approaches on Wino-X.}
\label{table:ours_pm_winox}
\end{table}
\begin{table}[ht!]
\centering
\resizebox{\columnwidth}{!}{
\begin{tabular}{lrrrrrr}
\toprule
&
  \multicolumn{1}{c}{de} &
  \multicolumn{1}{c}{es} &
  \multicolumn{1}{l}{fr} &
  \multicolumn{1}{l}{ru} &
  \multicolumn{1}{c}{he} &
  \multicolumn{1}{c}{ar} \\
\midrule
Opus-MT       & 66            & 60.4          & 56.5        & 50.2          & 56.6        & \textbf{59.8}          \\
Gender data     & 73.3          & 66.7          & \textbf{60.3}        & 52.4          & 60.9        & 59.7          \\
$+$coref markers  & \textbf{76.8}          & \textbf{67.9}          & 60.1        & \textbf{53.2}          & \textbf{63.5}        & 55.6          \\
\bottomrule
\end{tabular}}
\caption{Gender Accuracy results of our fine tuning approaches on WinoMT.}
\label{table:ours_acc_wino-mt}
\end{table}

\section{Related Work}

The study of coreference has a long tradition in machine translation.
A long line of work uses pronoun translation as a way of measuring coreference, since BLEU-based evaluation was shown to be insufficient for measuring improvement in coreference \citep{hardmeier-federico-2010-modelling}.

An alternative evaluation methodology is using automatic reference-based methods that produce a score based on word alignment between the source, reference translation, and translation output, and identification of pronouns in them, such as AutoPRF \citep{hardmeier-federico-2010-modelling} and APT~\citep{miculicich-werlen-popescu-belis-2017-validation}. 
Nevertheless, a later human meta-evaluation showed substantial disagreement between these metrics and human annotators, especially because of the existence of valid alternative translations and pronouns than the ones used in the reference \cite{guillou-hardmeier-2018-automatic}. 
Based on these conclusions, \citet{sennrich-2017-grammatical} developed a scoring-based evaluation approach that compares model scores of a predefined set of correct and incorrect translations and evaluates how often the model selects the correct option.

Our method extends \citep{stanovsky-etal-2019-evaluating}, which used a reference-free approach by aligning the source and candidate translation, but focused on entity translation accuracy to evaluate gender bias in MT models.
The availability of references was assumed by most previous work \citep{guillou-hardmeier-2016-protest, bawden-etal-2018-evaluating, muller-etal-2018-large, stojanovski-etal-2020-contracat, emelin-sennrich-2021-wino}, where most of them are limited to a single language pair. 
The flexibility afforded by a reference-free approach allows us to evaluate any target language for which an alignment model and morphological analyzer are available. Moreover, our approach is not restricted by a predefined set of translations and can also correctly detect valid translations that are different from the reference.

Several previous methods aimed to improve the coreference abilities of MT models and reduce undesirable biases, by modifying the training data in ways that share some similarities with our method. \citet{vanmassenhove-etal-2018-getting} incorporate a ``speaker gender'' tag into training data, allowing gender to be conveyed at the sentence level. Similarly, \citet{moryossef-etal-2019-filling} added a prefix to disambiguate the coreference in the sentence. 
\citet{stojanovski-fraser-2018-coreference} used oracle-based approach to inject new tokens indicating the pronoun translation and its gender into the source sentence. 
Our method is novel in the way it enriches the data with coreference signal using only the source-side signal, and thus requires only an English coreference resolution model without the need for coreference annotation in the target language.
\section{Conclusion}

Our work is the first to present an automatic methodology for assessing the coreference capabilities of MT models, that can be applied in any target language and does not require any target side annotations. Furthermore, to the best of our knowledge, we are the first to conduct a large-scale multilingual coreference evaluation study on prominent open-source and commercial MT models, and compare them against state-of-the-art coreference resolvers on three challenging benchmarks. Finally, based on the superior results of coreference resolvers, we propose a novel approach to improve the coreference capabilities of MT models, that outperforms or achieves comparable results to strong and larger MT models. Despite this substantial gain, there is still a performance gap between our model and state-of-the-art coreference resolvers. We hope that our work, and specifically our automatic evaluation methodology, will encourage future research to improve the coreference capabilities of MT models.

Future work can expand our approach to account for number and person agreement phenomena, investigate how to extend our approach to more coreference clusters and more mentions per cluster in intra-sentential as well as inter-sentential settings. Moreover, we intend to investigate how different morphological attributes affect MT models' coreference abilities.

\section*{Limitations}

Even though our study presents the first large-scale multilingual coreference evaluation study in MT, it still has some limitations that could be addressed in future work. First, our methodology provides an upper bound to the coreference capabilities based on detecting gender valuations. While this could allow for a  controlled evaluation experiment, this upper bound can become non-indicative in cases where gender assignment is not a discriminative factor. This can be addressed by accounting for more semantic and syntactic constraints that the translation needs to follow (e.g., singular/plural agreement). 

Second, our setting addresses one entity and a single co-referring pronoun in the naturalistic sentences experiment. Our methodology could in principle be augmented to deal with more coreference clusters and mentions per cluster. Another possible extension is to include event coreference in addition to entity coreference. For example, in this work, we focus only on the anaphoric function of the pronoun ``it'' but further research can also examine the event function of ``it'' \citep{loaiciga-etal-2017-disambiguating}.

Third, MT models should generally produce translations with accurate gender inflection for all words. However, in this work, we focus on the \textit{coreference} capabilities of MT models by evaluating gender agreement between coreferring \textit{entity} mentions. Future research can extend our evaluation methodology to assess the gender inflection of verb and adjective translation (e.g., the gender of ``big'' and ``small'' in Figure~\ref{fig:method_new}), using additional tools and resources such as a semantic role labeling model and a dependency parser.


Finally, although in Section \ref{sec:improving} we show big gains from the fine-tuning approach, it is clear that there is much room for improving the coreference capabilities of MT models, especially with regard to the performance of state-of-the-art coreference resolvers. We hope this work will help others develop MT models with better coreference capabilities.

\section*{Acknowledgements}

We thank the reviewers for their insightful comments and suggestions. This work was partially supported by the Israeli Ministry of Science and Technology (grant no. 2088). Arie Cattan is partially supported by the PBC fellowship for outstanding PhD candidates in data science.

\bibliography{anthology,custom}
\bibliographystyle{acl_natbib}

\newpage
\appendix

\section{Human Validation Results}
\label{sec:appendix:05_human_validation_results}

Table \ref{table:human_val_results} shows the complete human annotations results. The results indicate that alignment and gender prediction are accurate in most languages. In Arabic and Hebrew, the alignment error occurs more. A possible explanation for that can be the fact that both those languages are partial pro-drop languages. To verify that those results will not affect our measurement, we verified that the error has similar consistency distributions as the rest of our results.  

\begin{table}[t]
\centering
\resizebox{\columnwidth}{!}{
\begin{tabular}{lrrrrrr}
\toprule
 & \multicolumn{1}{l}{he} & \multicolumn{1}{l}{ar} & \multicolumn{1}{l}{es} & \multicolumn{1}{l}{ru} & \multicolumn{1}{l}{fr} & \multicolumn{1}{l}{de} \\
\midrule
alignment error         & 14  & 19  & 4   & 4   & 4   & 0   \\
gender prediction error & 7   & 7   & 0   & 7   & 2   & 2   \\
correct annotation       & 79  & 74  & 96  & 139 & 144 & 148 \\
Total $\#$ of annotation        & 100 & 100 & 100 & 150 & 150 & 150 \\
\bottomrule
\end{tabular}}
\caption{Human validation results on our three evaluation datasets and six target languages.}
\label{table:human_val_results}
\end{table}

\section{BUG Consistency Results}
\label{sec:appendix:03_bug_results}

\begin{table*}[ht!]
\centering
\begin{tabular}{lrrrrrr}
\toprule
&
  \multicolumn{1}{c}{en$\rightarrow$de} &
  \multicolumn{1}{c}{en$\rightarrow$es} &
  \multicolumn{1}{l}{en$\rightarrow$fr} &
  \multicolumn{1}{l}{en$\rightarrow$ru} &
  \multicolumn{1}{c}{en$\rightarrow$he} &
  \multicolumn{1}{c}{en$\rightarrow$ar} \\
\midrule

Opus-MT         & 79.8 & 85.1 & 86.2 & 67.3 & 88.8 & 88.0   \\
gender data     & 84.1 & \textbf{87.4} & 88.7 & 69.9 & \textbf{90.2} & 91.3 \\
$+$coref markers  & \textbf{84.2} & \textbf{87.4} & \textbf{88.8} & \textbf{70.1} & 89.8 & \textbf{91.8} \\

\bottomrule
\end{tabular}
\caption{Target-side consistency results of our implicit and explicit fine tuning approaches on BUG.}
\label{table:ours_cons_BUG}
\end{table*}

Table \ref{table:ours_cons_BUG} presents the target-side consistency results of the Opus-MT baseline model and all our fine tuning approach on BUG. Similarly to Wino-X and WinoMT, our fine-tuned models outperform the baseline.

\section{Omitted Cases}
\label{sec:appendix:04_models_exampels}

\begin{table*}[ht]
\centering
\resizebox{1\textwidth}{!}{
\begin{tabular}{p{5cm}p{5.5cm}p{5cm}}
\toprule
Source & {[}Target lang.{]} Predicted translation & Phenomenon \\
  \midrule
\parbox{5cm}{The developer argued with the designer because \textcolor{blue}{his} idea cannot be implemented.} &
  \parbox{5.5cm}{{[}FR.{]} Le développeur a argumenté avec le concepteur parce que \textcolor{blue}{son} \textcolor{blue}{idée} ne peut pas être mis en œuvre.} &
  \parbox{5cm}{  ``son'' is male because the possessed noun (``idée'') is male.} \\
  \midrule
\parbox{5cm}{The doctor asked the nurse to help \textcolor{red}{her} in the procedure} &
  \parbox{5.5cm}{{[}ES.{]} El doctor le pidio a la enfermera que le ayudara con el procedimiento} &
  \parbox{5cm}{In Spanish, the pronoun ``her'' is dropped in the translation.} \\
  \bottomrule

  
\end{tabular}
}
\caption{Examples of omitted sentences from our evaluation datasets and their translations. Words in blue and red indicate male and female entities, respectively.  }
\label{table:lang_exampels_app}
\end{table*}

Table \ref{table:lang_exampels_app} shows translation examples from English to French and Spanish that demonstrate unique features in each language. 
The first example shows a French translation that contains a possessive pronoun, which does not indicate the gender of the possessor. The second example shows a Spanish translation where the pronoun is omitted. In both cases, we can obtain a correct translation without information concerning the aligned pronoun gender, we therefore exclude them from the evaluation.




\section{Pronoun Translation Accuracy}
\label{sec:appendix:01_gmf_pm}

\begin{table}[tb!]
\begin{tabular}{llll}
\toprule
               & German         & French         & Russian        \\
\midrule
mBART50   & 38.1          & \textbf{47.2} & 34.7          \\
M2M00\_418M & 33.8          & 39.5          & 32.4          \\
M2M100\_1.2B & 38.5          & 39.0          & 33.7          \\
Google         & \textbf{44.2}          & 34.8          & 35.8          \\
Microsoft      & 43.1          & 35.7          & \textbf{37.8}           \\
EasyNMT       & 39.9          & 31.8          & 37.0             \\
\bottomrule
\end{tabular}
\caption{Pronoun accuracy results of commercial and open-source MT models on Wino-X.}
\label{table:gmf_pm_winox}
\end{table}

Table \ref{table:gmf_pm_winox} shows pronoun accuracy results of our baselines on Wino-X. We can notice those results are similar to our consistency results although our methodology does not use any annotated target data. Moreover, those results clearly show that MT models struggle with translating sentences that demand solving the source side coreference resolution.

\section{Computing Infrastructure}
\label{sec:appendix:08_comp_rec}

We fine tuned our models using 4 NVIDIA GTX Titan Black GPUs. The run time of the models varies between one hour to 24 hours depending on dataset size.

\end{document}